\documentclass[10pt,twocolumn,letterpaper]{article}

\usepackage{iccv}
\usepackage{times}
\usepackage{epsfig}
\usepackage{graphicx}
\usepackage{amsmath}
\usepackage{amssymb}

\usepackage{booktabs}
\usepackage{multirow} 
\usepackage{colortbl}
\usepackage{subfig}
\usepackage{bbding}
\usepackage{algorithm}
\usepackage{algorithmic}

\usepackage[pagebackref=true,breaklinks=true,letterpaper=true,colorlinks,bookmarks=false]{hyperref}

\iccvfinalcopy 

\def\iccvPaperID{2492} 

\ificcvfinal\fi

\begin{document}
\def\X{\textcolor{red}{\XSolidBrush}}
\def\C{\textcolor{green}{\CheckmarkBold}}
\def\b{\textcolor{blue}}
\def\textb{\textcolor[gray]{.4}}

\title{ Gloss-free Sign Language Translation: Improving from Visual-Language Pretraining
}

\author{Benjia Zhou$^{1}$,  Zhigang Chen$^{2,3}$\thanks{Benjia Zhou and Zhigang Chen contributed equally to this paper.},
Albert Clapés$^{4,5}$, Jun Wan$^{1,2,3}$\thanks{Corresponding author.} ,  Yanyan Liang$^{1}$,  \\
Sergio Escalera$^{4,5,6}$,  Zhen Lei$^{2,3,7}$,  Du Zhang$^{1}$ \\
$^{1}$MUST, Macau, China;
$^{2}$UCAS, China;
$^{3}$MAIS, CASIA, China;
$^{4}$Universitat de Barcelona, Spain; \\
$^{5}$Computer Vision Center, Spain;
$^{6}$AAU, Aalborg, Denmark;
$^{7}$CAIR, HKISI, CAS, Hong Kong, China
}

\maketitle
\ificcvfinal\thispagestyle{empty}\fi

\begin{abstract}
    Sign Language Translation (SLT) is a challenging task due to its cross-domain nature, involving the translation of visual-gestural language to text. Many previous methods employ an intermediate representation, \ie, gloss sequences, to facilitate SLT, thus transforming it into a two-stage task of sign language recognition (SLR) followed by sign language translation (SLT).
    However, the scarcity of gloss-annotated sign language data, combined with the information bottleneck in the mid-level gloss representation, has hindered the further development of the SLT task.
    To address this challenge, we propose a novel \textbf{G}loss-\textbf{F}ree \textbf{SLT} based on \textbf{V}isual-\textbf{L}anguage \textbf{P}retraining (\textbf{GFSLT-VLP}), which improves SLT by inheriting language-oriented prior knowledge from pre-trained models, without any gloss annotation assistance. Our approach involves two stages: (i) integrating Contrastive Language-Image Pre-training (CLIP) with masked self-supervised learning to create pre-tasks that bridge the semantic gap between visual and textual representations and restore masked sentences, and (ii) constructing an end-to-end architecture with an encoder-decoder-like structure that inherits the parameters of the pre-trained Visual Encoder and Text Decoder from the first stage. The seamless combination of these novel designs forms a robust sign language representation and significantly improves gloss-free sign language translation.
    In particular, we have achieved unprecedented improvements in terms of BLEU-4 score on the PHOENIX14T dataset ($\geq$+5) and the CSL-Daily dataset ($\geq$+3) compared to state-of-the-art gloss-free SLT methods. Furthermore, our approach also achieves competitive results on the PHOENIX14T dataset when compared with most of the gloss-based methods\footnote{\url{https://github.com/zhoubenjia/GFSLT-VLP}}.
\end{abstract}

\begin{figure}[!t]
\centering
\subfloat[Gloss-based approach.]{\includegraphics[width=1.0\linewidth]{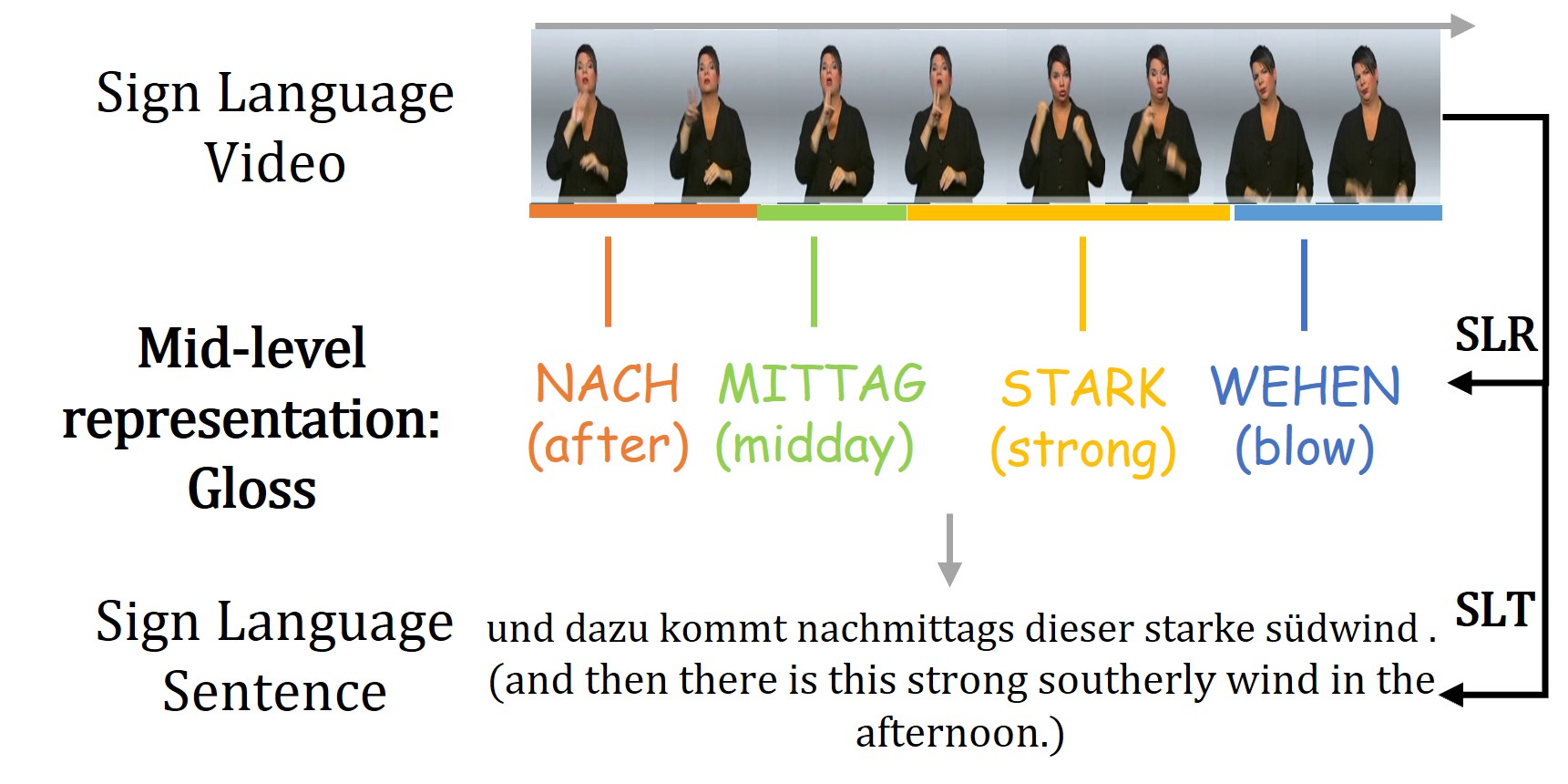}%
\label{subfig:gloss_based}}
\hfil
\subfloat[Gloss-free approach (ours).]{\includegraphics[width=1.0\linewidth]{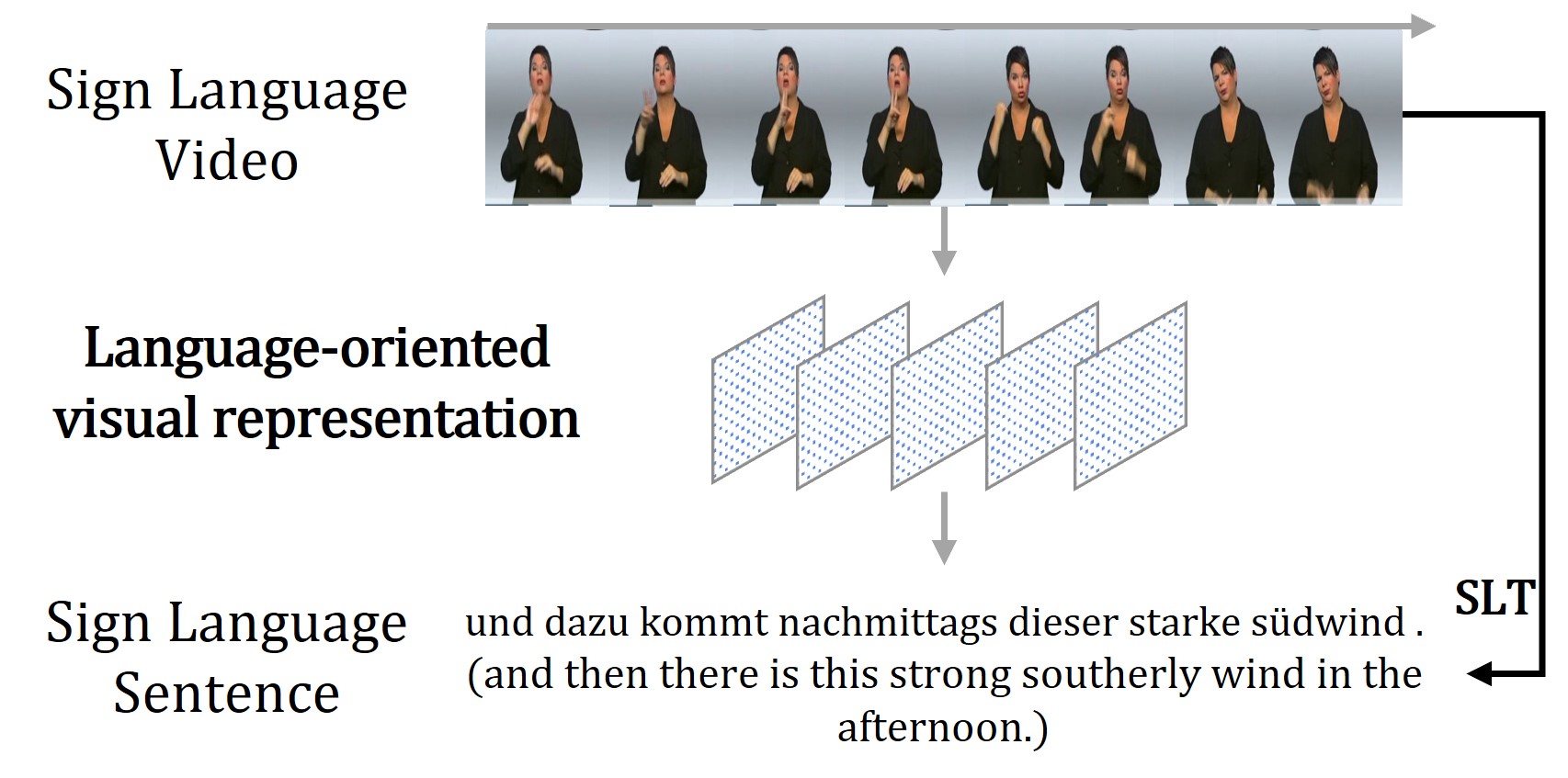}%
\label{subfig:gloss_free}}
\caption{Two SLT approaches: (a) using gloss sequences as intermediate representations, \eg, Sign2Gloss2Text (directly), Sign2Text (indirectly), (b) not using gloss info throughout the training/inference process.
}
\label{fig:approach}
\end{figure}
\section{Introduction}
\label{sec:intro}
Sign language is the main medium of communication among deaf people. To facilitate effective communication with hard-of-hearing people, developing Sign Language Translation (SLT) techniques is a promising direction. SLT refers to translating sign language into fluent spoken language sentences, which is more challenging than traditional Natural Machine Translation (NMT) due to its cross-domain translation nature and the scarcity of annotated data.

Recently, a growing body of literature~\cite{camgoz2020sign, zhou2021improving, hao2021self, chen2022twostream, chen2022simple} has promoted the SLT by directly or indirectly employing the intermediate representations, namely sign glosses. Gloss is a simplified representation of each sign language in continuous video
as illustrated in Figure \ref{subfig:gloss_based}.
Although gloss-based methods have significantly improved the SLT performance compared to end-to-end gloss-free approaches (as illustrated in Figure \ref{subfig:gloss_free}), the former still suffers from the following problems:
(i) annotating glosses is a labor-intensive task, which requires fine-grained alignment and labeled by specialists, significantly constraining the scalability of gloss-based SLT methods
and (ii) the gloss-based approach introduces an information bottleneck in the mid-level gloss representation~\cite{camgoz2020sign}, which limits the network’s ability to understand sign language as the translation model can only be as good as the sign gloss annotations it was trained from.

Inspired by CLIP~\cite{radford2021learning}, which utilizes natural language supervision for image representation learning, we discovered that learning language-indicated visual representation from sign language videos is an effective pre-training task for SLT as it establishes a potential connection between visual signs and language context.
However, directly transferring CLIP to SLT is not advisable due to two reasons: (i) it cannot perform joint pretraining of the Visual Encoder and Text Decoder for SLT, and (ii) sufficient SLT data is required to support this pretraining task. To address these challenges, we need to tackle two critical questions: (i) how to achieve efficient joint pretraining on the limited SLT dataset? and (ii) how to ensure that the pretraining model offers the most effective assistance for the downstream SLT task?

\begin{figure*}[!htp]
  \centering
  \includegraphics[width=1.0\linewidth]{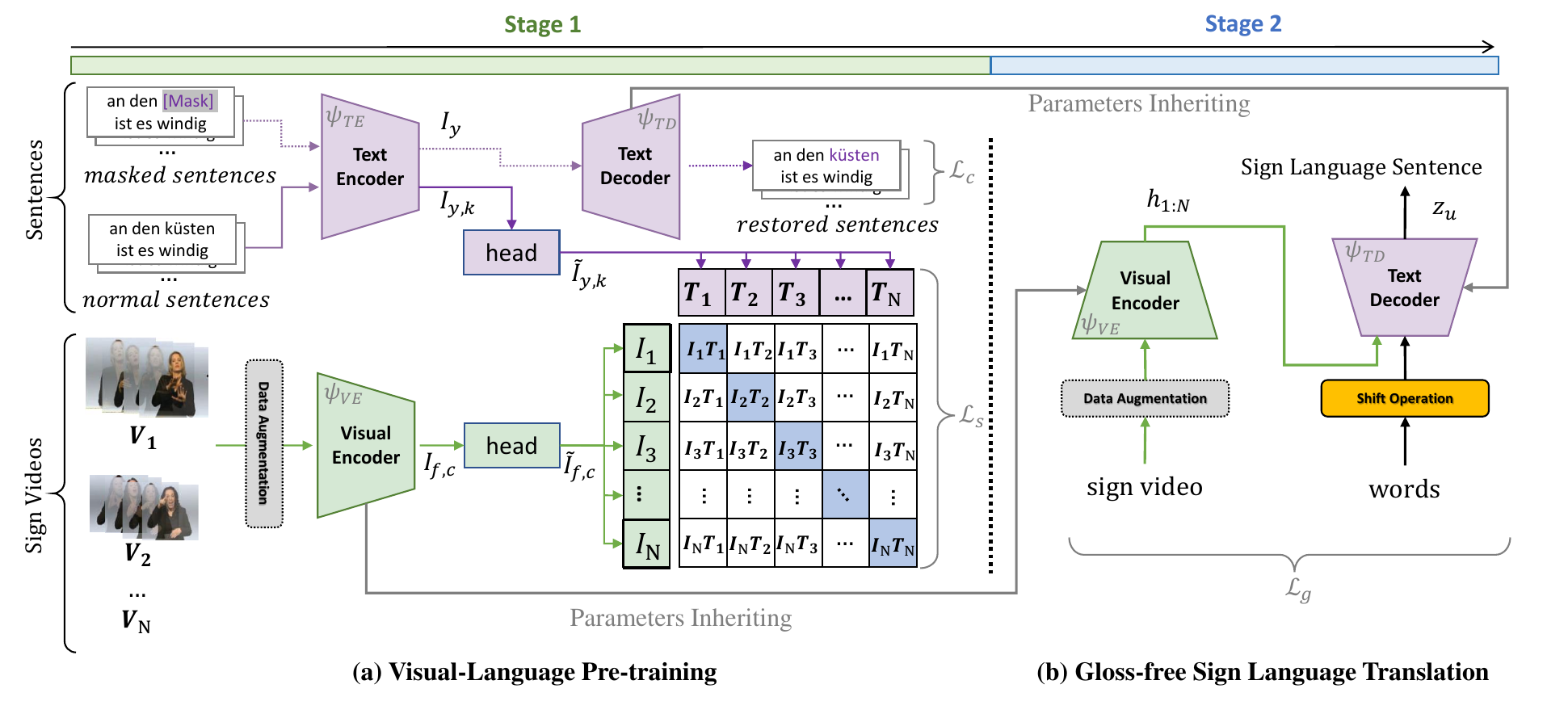}

  \caption{\textbf{Method Overview.} GFSLT-VLP improves the SLT by (a) performing Visual-Language Pretraining in stage 1 first, and then (b) transferring parameters of the pre-trained Visual Encoder and Textual Decoder in stage 2. Wherein $\mathrm{N}$ indicates the number of samples in a mini-batch.
  }
  \label{fig:pipeline}
\end{figure*}

To address the first challenge, we propose a solution at both the algorithmic and data levels. At the algorithmic level, we introduce a novel pre-training paradigm, called VLP (Visual-Language Pretraining), which incorporates the masked self-supervised learning paradigm together with CLIP as illustrated in Figure \ref{fig:pipeline}\textcolor{red}{a}. Specifically, we design a \textit{pretext task} that aligns visual and textual representations in a joint multimodal semantic space, guiding the Visual Encoder to learn language-indicated visual representations. Meanwhile, we introduce masked self-supervised learning into the pre-training mechanism to guide the Text Decoder to capture the syntactic and semantic properties of sign language sentences.
At the data level, we investigate a set of strong data augmentation techniques for sign videos to increase the diversity of visual data. This is an aspect that has not always been adequately addressed in previous SLT methods.

To cope with the second aspect, as shown in Figure \ref{fig:pipeline}\textcolor{red}{b}, we construct an end-to-end \textbf{G}loss-\textbf{F}ree \textbf{SLT} architecture with an encoder-decoder-like structure called GFSLT, which inherits the parameters of the pre-trained Visual Encoder and Text Decoder from the first stage. This architecture enables us to directly encode visual representations into spoken sentences without requiring any intermediate projection or supervision. Moreover, unlike other methods~\cite{camgoz2020sign, zhou2021spatial, chen2022simple} that only fine-tune the spatial feature extractor (visual embedding module) in the Visual Encoder, we fine-tune both the spatial feature extractor and the temporal relationship modeling network (Transformer encoder) as a unified whole.

In summary, the main contributions are listed:
\begin{itemize}
    \item In this work, we have achieved unprecedented improvements in the BLEU-4 score for SLT without using gloss annotations. Specifically, compared with state-of-the-art gloss-free SLT methods, our method has got $\geq$+5 and $\geq$+3 improvements on the PHOENIX14T dataset and CSL-Daily dataset, respectively. We believe that these improvements represent a significant breakthrough in the task of gloss-free SLT.
    \item To the best of our knowledge, this is the first attempt to introduce the VLP strategy to align visual and textual representations in a joint semantic space in the gloss-free SLT task.
    \item We propose a novel pre-training paradigm that incorporates masked self-supervised learning together with contrastive language-image pre-training to facilitate the gloss-free SLT task. This approach represents a significant improvement over previous methods and has the potential to greatly enhance the accuracy and efficiency of SLT systems.
\end{itemize}






\section{Related Works}
Generally speaking, there are two methods for Sign Language Translation (SLT), namely, gloss-based and gloss-free. Before briefly surveying works along these two directions, we first introduce the Sign Language Recognition (SLR) task as it is an essential step for gloss-based SLT methods.
\subsection{Sign Language Recognition}
Sign Language Recognition (SLR) consists of two different tasks: Isolated Sign Language Recognition (ISLR) and Continuous Sign Language Recognition (CSLR). The goal of ISLR is to translate an isolated sign into a corresponding single sign language word \cite{imashev2020dataset,joze2018ms,li2020word,li2020transferring}, which is somewhat similar to the isolated gesture recognition task \cite{junwan2013JMLR,konevcny2014one, junwan2016pami, Zhou_Li_Wan_2021, Zhou_2022_CVPR, zhou2023pami, yu2021searching}. CSLR is a more challenging task, which is dedicated to recognizing a continuous video of sign language into ordered sign language words, referred to as gloss sequences~\cite{camgoz2020sign,chen2022twostream,hao2021self,hu2022temporal,koller2019weakly,min2021visual,zhou2021spatial,zuo2022c2slr}. Previous SLT work often utilized CSLR as a pre-task to predict gloss or obtain better visual representations~\cite{camgoz2020sign,zhou2021spatial,zhou2021improving,chen2022simple,chen2022twostream}. Such methods often have high requirements on the accuracy of CSLR. In this work, however, we abandon gloss sequences entirely and explore a new gloss-free SLT approach.
\subsection{Gloss-based Sign Language Translation}
To improve the Sign Language Translation (SLT), several works have employed the mid-level representation of sign glosses.
SLRT~\cite{camgoz2020sign} first introduces a Transformer-based encoder-decoder framework to perform end-to-end SLT. This approach improves performance by using a Connectionist Temporal Classification (CTC) loss to soft-match sign representations and gloss sequences.
STMC-T\cite{zhou2021spatial} approaches sign language understanding with multi-cue learning. It models sequence information by introducing intra-cue and inter-cue CTC~\cite{graves2006connectionist} losses. SignBack\cite{zhou2021improving} attempts to introduce advanced machine translation techniques such as back-translation~\cite{zhou2021improving} into SLT. Moreover, thanks to the successful application of transfer learning on NMT, Chen \etal ~\cite{chen2022simple, chen2022twostream} made the first attempt to introduce large language models into SLT.
All the above methods used the Gloss annotation directly or indirectly in SLT model training. However, in our case, we completely abandon the Gloss annotation because its existence limits the scale of sign language datasets. Instead, a more general design of sign language pre-training is introduced in this paper.

\subsection{Gloss-free Sign Language Translation}
Gloss-free SLT refers to the absence of gloss supervision throughout the training and testing stages, including pre-training and fine-tuning. NSLT \cite{camgoz2018neural} utilized CNN+RNN to model SLT end-to-end, where CNN learned visual features of sign language, and RNN with attention mechanism \cite{bahdanau2015neural,luong2015effective} performed sequence learning and text modeling. TSPNet \cite{li2020tspnet} designed inter-scale attention and intra-scale attention to model local and global contextual semantic information of sign language video clips for better visual feature learning. In contrast, CSGCR \cite{zhao2021conditional} aimed to improve the accuracy and fluency of SLT by proposing three modules: word existence verification, conditional sentence generation, and cross-modal re-ranking to learn better grammatical features. However, aligning the two modalities without gloss supervision is challenging due to the significant difference in the order of sign language videos and spoken language. Consequently, the performance of gloss-free SLT is much lower than that of gloss-based SLT.
In this paper, we adopt a VLP-based strategy to obtain better cross-modal representations and significantly narrow the performance gap between gloss-free SLT and gloss-based SLT.

\begin{figure*}[!htp]
\centering
\subfloat[GFSLT Model]{\includegraphics[width=0.5\linewidth]{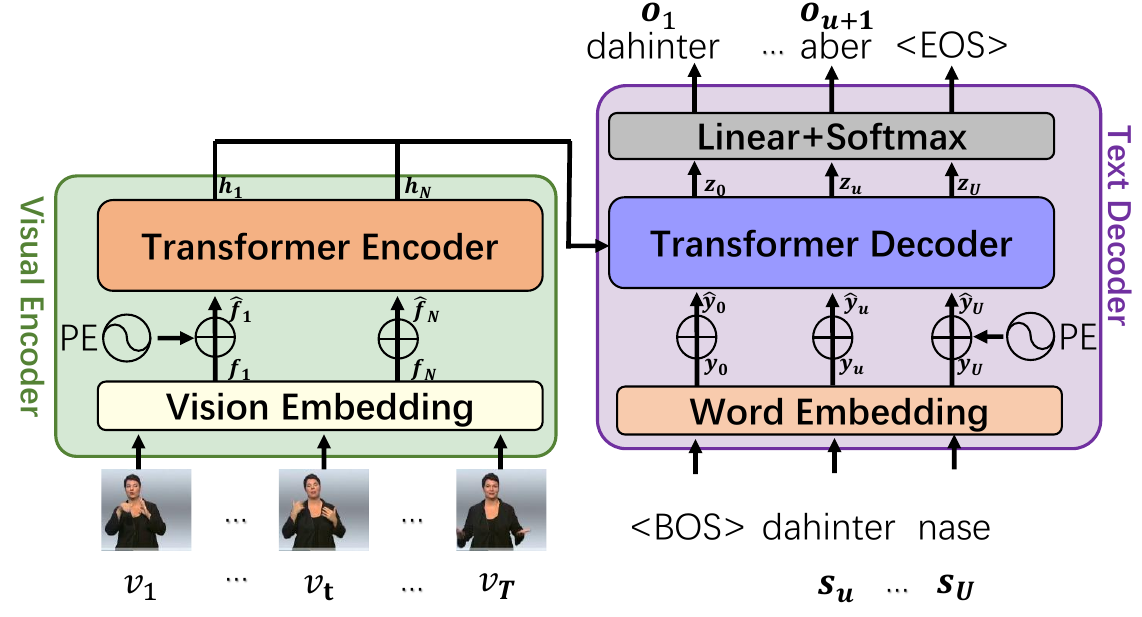}%
\label{fig:transformer}}
\hfil
\subfloat[Vision Embedding]{\includegraphics[width=0.4\linewidth,height=0.26\linewidth]{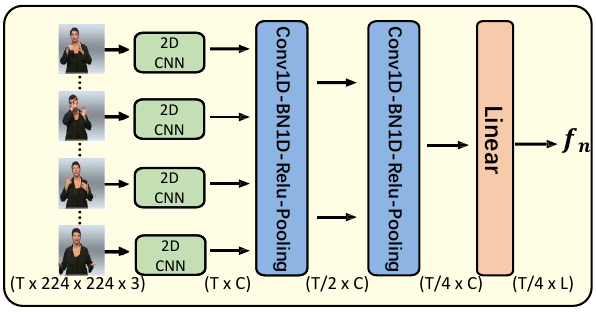}%
\label{fig:vision embedding}}
\caption{(a) The framework of the Gloss-free SLT Model, where PE means Positional Encoding. (b) The structure of the Vision Embedding layer.}
\label{fig:slt}
\end{figure*}
\section{Method}

In this paper, we suggest that language-indicated visual representations enjoy both low-redundancy and high-abstract properties of language information that can improve SLT.
To this end, we introduce a new pre-training paradigm for SLT that combines masked self-supervised learning with CLIP, allowing us to jointly pre-train the Visual Encoder $\psi_{VE}(\cdot)$ and Text Decoder $\psi_{TD}(\cdot)$ for the downstream GFSLT model (Section \ref{sec:pretrain}).
Subsequently, we transfer the parameters of the pre-trained Visual Encoder $\psi_{VE}^*(\cdot)$ and Text Decoder $\psi_{TD}^*(\cdot)$ to the GFSLT model $\psi_{GFSLT}(\cdot)$ meticulously to enhance its translation capabilities (Section \ref{sec:finetune}). Algorithm \ref{alg:Method} elaborates our entire algorithm flow.
\begin{algorithm}[!htp]
    \caption{Two-Stage Gloss-free SLT.}
    \label{alg:Method}
    \textit{{\bfseries Stage1:} Visual-Language Pre-training (VLP)}

    \begin{algorithmic}[1]
        \STATE \textbf{Input:} Dataset $\mathcal{D} =\{V^{(n)},S^{(n)}\}_{n=1}^N$
        \STATE \small{Initialize the parameters $\Theta_*$ of $\psi_{VE}(\cdot)$, $\psi_{TD}(\cdot)$ and $\psi_{TE}(\cdot)$}
        \WHILE{not converged}
            \FOR{$V^{(i)},S^{(i)}$ in $\mathcal{D}$}
                \STATE \small{Update the $\psi_{VE}(\cdot)$ by descending $\bigtriangledown \mathcal{L}_s(\Theta_{VE}, V^{(i)}) + \mathcal{L}_s(\Theta_{TE}, S^{(i)})$}
                \STATE \small {Obtain the masked sentences $\Tilde{S}^{(i)}$}
                \STATE  \small{Update the $\psi_{TD}(\cdot)$ by descending $\bigtriangledown \mathcal{L}_c(\Theta_{TD}, \Tilde{S}^{(i)})$}
            \ENDFOR
        \ENDWHILE
        \STATE \textbf{Output:} $\psi_{VE}^*(\cdot)$ and $\psi_{TD}^*(\cdot)$
    \end{algorithmic}
    \textit{{\bfseries Stage2:} Gloss-Free Sign Language Translation (GFSLT)}

    \begin{algorithmic}[1]
        \STATE \small{Initialize the parameters $\Theta_{GFSLT}$ of $\psi_{GFSLT}(\cdot)$ with $\psi_{VE}^*(\cdot)$, $\psi_{TD}^*(\cdot)$}
        \WHILE{not converged}
            \FOR{$V^{(i)},S^{(i)}$ in $\mathcal{D}$}
                \STATE  \small{Update the $\psi_{GFSLT}(\cdot)$ by descending

                $\bigtriangledown \mathcal{L}_g(\Theta_{GFSLT}, \Tilde{S}^{(i)})$}
            \ENDFOR
        \ENDWHILE
        \STATE \textbf{Output:} $\psi_{GFSLT}^*(\cdot)$
    \end{algorithmic}
\end{algorithm}



\subsection{Visual-Language Pretraining}\label{sec:pretrain}
In order to learn language-indicated visual representations from sign videos, two crucial issues need to be considered: (i) \textbf{how to design a \textit{pretext task} that can effectively reduce the semantic gap between visual and textual representations?} and (ii) \textbf{how to achieve jointly pretraining on the limited SLT dataset?}


To cope with the first issue,  we draw inspiration from CLIP~\cite{radford2021learning} in the field of zero-shot transfer learning, which developed the ``image-to-text'' as a standardized input-output interface, allowing for transferable visual models from natural language supervision. CLIP~\cite{radford2021learning} has highlighted the advantages of learning from natural language over other task-agnostic pretraining methods, making it particularly suitable for sign language translation tasks.
In other words, learning visual representations through language supervision is a straightforward yet effective pretext task for SLT, given that SLT data inherently has an image-text pair structure.
From this insight, we present a new Visual-Language Pre-training scheme, termed VLP, as illustrated in Figure \ref{fig:pipeline}\textcolor{red}{a}. It trains a Visual Encoder $\psi_{VE}(\cdot)$ and a Text Encoder $\psi_{TE}(\cdot)$ jointly to predict the correct pairings of a batch of (sign video, language sentence) training examples. Formally, a video-text pair is first input into $\psi_{VE}(\cdot)$ and $\psi_{TE}(\cdot)$ in parallel to obtain corresponding high-dimensional semantic features:
\begin{equation}
    \begin{split}
       & I_f = \psi_{VE}(V), \quad V=(v_1,...,v_T); \\
        & I_y = \psi_{TE}(S), \quad S=(s_1,...s_U);
    \end{split}
\end{equation}
where $V$ is a sign language video with $T$ frames and $S$ is a spoken language sentence with $U$ words.




\vspace{+0.1cm}
\noindent\textbf{Visual Encoder:} The Visual Encoder consists of a Visual Embedding layer, as shown in Figure \ref{fig:vision embedding}, followed by a Transformer Encoder with multiple layers.
Each frame of the video is first encoded by the weight-sharing 2D CNN layers;
The resulting visual encoding is then fed through two temporal blocks, which use a combination of Conv1D-BN-Relu-Maxpooling to capture short-term dependencies. Finally, the features are passed through the Transformer Encoder to capture long-term dependencies in the video.

\vspace{+0.1cm}
\noindent\textbf{Text Encoder:} To encode text data effectively, it is crucial to have a strong Text Encoder. As a result, we have opted to use the parameters initialized encoder with 12 layers in mBART~\cite{liu2020multilingual}. This is an NMT model that has been pre-trained on CC25~\cite{liu2020multilingual}, a multilingual corpus that covers 25 languages.

The captured visual features $I_f$ and textual features $I_y$ are then input to the corresponding heads to be linearly projected to the joint multimodal semantic space for similarity computation. Here, both heads are composed of a simple Linear layer. Formally, we can express the process as follows:
\begin{equation}
     \Tilde{I}_{f,c} = \mathrm{Linear}(I_{f,c}),
    \Tilde{I}_{y,k} = \mathrm{Linear}(I_{y,k});
\end{equation}
where $I_{f,c}$ denotes the activation of the last layer of the Visual Encoder at the [CLS] token\footnote{
It is a special token that is added to make a global representation of the whole sequence.
}, and $I_{y,k}$ denotes the activation of the last layer of the Text Encoder at the \textless EOS\textgreater  token\footnote{The text sequence is bracketed with \textless BOS\textgreater and \textless EOS\textgreater  tokens.}. Subsequently, similar to CLIP~\cite{radford2021learning}, $\Tilde{I}_{f,c}$ and $\Tilde{I}_{y,k}$ are layer-normalized and pairwise-scaled, and then used to calculate the loss value via a symmetric Cross-entropy loss function:
\begin{equation}
    \mathcal{L}_s = -\frac{1}{2}\big(\sum{V \log(\Tilde{I}_{f,c})} + \sum{ S \log(\Tilde{I}_{y,k})}\big)
\end{equation}


To tackle the second question, we take a dual approach at both the algorithm and data levels.
At the algorithm level, as illustrated in Figure \ref{fig:pipeline}\textcolor{red}{a}, we combine two pre-training paradigms - masked self-supervised learning and visual-language supervision learning - to achieve end-to-end joint pre-training.
This consideration stems from the fact that different pre-training paradigms can capture different aspects of the data, and combining them can provide a more comprehensive representation of the data \cite{kendall2018multi, kornblith2019better}.
Let $\psi_{TD}(\cdot)$ denote the\textbf{ Text Decoder}, and its optimization goal is to restore the masked words in the input sentence,
which can be formulated as follows:
\begin{equation}
\min_{\Theta} \frac{1}{n} \sum_{i=1}^n \mathcal{L}_c\big(\psi_{TD}(\psi_{TE}^*(\Tilde{S}^{(i)})), S^{(i)}\big) 
\end{equation}
where $\Tilde{S}$ denotes the masked sentences, $n$ is the number of training samples, $\mathcal{L}_c$ is the loss function.
At the data level, we introduce strong data augmentation (implemented by VIDAUG library~\cite{vidaug}) for input videos in SLT, including geometric transformation, color space transformation, and temporal transformation. During training, we randomly combine these three augmentation methods to enlarge the data space.

In fact, this paradigm facilitates the Visual Encoder to acquire potent language representation skills that are similar to those of the Text Encoder, resulting in the generation of more robust and representative visual features. This is the reason why acquiring language-indicated visual features for SLT is feasible. Hence, after the Visual Encoder and Text Decoder have established such modeling capability, we employ them to perform the SLT task in the second stage.

\begin{table*}[!htp]
  \centering
  \begin{tabular}{@{}l|ccccc|ccccc@{}}
    \toprule
    \multirow{2}{*}{\textbf{Method}} &  \multicolumn{5}{c|}{\textbf{Dev}} &  \multicolumn{5}{c}{\textbf{Test}} \\
   \cline{2-11} & \textbf{B1} & \textbf{B2} & \textbf{B3} & \textbf{B4} &  \textbf{ROUGE} & \textbf{B1} & \textbf{B2} & \textbf{B3} & \textbf{B4} & \textbf{ROUGE}\\
    \midrule
    \rowcolor[gray]{.8} \multicolumn{11}{c}{Gloss-based} \\
    \midrule
    SLRT \cite{camgoz2020sign} & 47.26 & 34.40 & 27.05 & 22.38 & - & 46.61 & 33.73 & 26.19 & 21.32 & -\\
    STN-SLT~\cite{voskou2021stochastic} & 49.12& 36.29& 28.34& 23.23& -& 48.61 & 35.97 & 28.37 & 23.65 &-\\
    STMC-T \cite{zhou2021spatial} & 47.60 & 36.43 & 29.18 & 24.09 & 48.24 & 46.98 & 36.09 & 28.70 & 23.65 & 46.65\\
    BN-TIN-Transf.+SignBT \cite{zhou2021improving} & 51.11 & 37.90 & 29.80 & 24.45 & 50.29 & 50.80 & 37.75 & 29.72 & 24.32 & 49.54\\
    MMTLB \cite{chen2022simple} & 53.95 & 41.12 & 33.14 & 27.61 & 53.10 & 53.97 & 41.75 & 33.84 & 28.39 & 52.65\\
    TS-SLT \cite{chen2022twostream}& \textbf{54.32} & \textbf{41.99} & \textbf{34.15} & \textbf{28.66} & \textbf{54.08} & \textbf{54.90} & \textbf{42.43} & \textbf{34.46} & \textbf{28.95} & \textbf{53.48}\\
    \midrule
     \rowcolor[gray]{.8} \multicolumn{11}{c}{Gloss-free} \\
    \midrule
    NSLT \cite{camgoz2018neural} & 28.10 & 16.81 & 11.82 & 9.12 & 31.00 & 27.10 & 15.61 & 10.82 & 8.35 & 29.70 \\
    NSLT+Bahdanau \cite{camgoz2018neural,bahdanau2015neural} & 31.87 & 19.11 & 13.16 & 9.94 & 31.80 & 32.24 & 19.03 & 12.83 & 9.58 & 31.80 \\
    NSLT+Luong \cite{camgoz2018neural,luong2015effective} & 31.58 & 18.98 & 13.22 & 10.00 & 32.60 & 29.86 & 17.52 & 11.96 & 9.00 & 30.70 \\
    TSPNet \cite{li2020tspnet} & - & - & - & - & - & 36.10 & 23.12 & 16.88 & 13.41 & 34.96 \\
    CSGCR \cite{zhao2021conditional} & 35.85 & 24.77 & 18.65 & 15.08 & 38.96 & 36.71 & 25.40 & 18.86 & 15.18 & 38.85 \\
    GASLT~\cite{yin2023gloss}   & - & - & - & -& - & 39.07  & 26.74 & 21.86 & 15.74 & 39.86\\
    \midrule
     \textbf{GFSLT (ours)} &  {41.97} &  {31.04} &  {24.30} &  {19.84} &  {40.70} &  {41.39} &  {31.00} &  {24.20} &  {19.66} &  {40.93} \\
    \textbf{GFSLT-VLP (ours)} & \textbf{44.08} & \textbf{33.56} & \textbf{26.74} & \textbf{22.12} & \textbf{43.72} & \textbf{43.71} & \textbf{33.18} & \textbf{26.11} & \textbf{21.44} & \textbf{42.49}\\
    \midrule
    \textcolor[gray]{.4}{Improvement} & \textcolor[gray]{.4}{+8.23} & \textcolor[gray]{.4}{+8.79} & \textcolor[gray]{.4}{+8.09} & \textcolor[gray]{.4}{+7.04} & \textcolor[gray]{.4}{+4.76} & \textcolor[gray]{.4}{+4.64} & \textcolor[gray]{.4}{+6.44} & \textcolor[gray]{.4}{+4.25} &\textcolor[gray]{.4}{+5.70} & \textcolor[gray]{.4}{+2.63}\\
    \bottomrule
  \end{tabular}
  \caption{Experimental results on PHOENIX14T dataset. We report BLEU-n in B-n columns and ROUGE. \textcolor[gray]{.4}{Improvement} represents the result of comparison with the latest gloss-free methods. }
  \label{tab:PHOENIX14T}
\end{table*}
\begin{table*}[ht]
  \centering
  \begin{tabular}{@{}l|ccccc|ccccc@{}}
    \toprule
    \multirow{2}{*}{\textbf{Method}} &  \multicolumn{5}{c|}{\textbf{Dev}} &  \multicolumn{5}{c}{\textbf{Test}} \\
   \cline{2-11} & \textbf{B1} & \textbf{B2} & \textbf{B3} & \textbf{B4} &  \textbf{ROUGE} & \textbf{B1} & \textbf{B2} & \textbf{B3} & \textbf{B4} & \textbf{ROUGE}\\
    \midrule
    \rowcolor[gray]{.8} \multicolumn{11}{c}{Gloss-based} \\
    \midrule
    SLRT \cite{camgoz2020sign} & 37.47 & 24.67 & 16.86 & 11.88 & 37.96 & 37.38 & 24.36 & 16.55 & 11.79 & 36.74\\
    BN-TIN-Transf. \cite{zhou2021improving} & 40.66 & 26.56 & 18.06 & 12.73 & 37.29 & 40.74 & 26.96 & 18.48 & 13.19 & 37.67\\
    BN-TIN-Transf.+SignBT \cite{zhou2021improving} & 51.46 & 37.23 & 27.51 & 20.80 & 49.49 & 51.42 & 37.26 & 27.76 & 21.34 & 49.31\\
    MMTLB \cite{chen2022simple} & 53.81 & 40.84 & 31.29 & 24.42 & 53.38 & 53.31 & 40.41 & 30.87 & 23.92 & 53.25\\
    TS-SLT \cite{chen2022twostream}& \textbf{55.21} & \textbf{42.31} & \textbf{32.71} & \textbf{25.76} & \textbf{55.10} & \textbf{55.44} & \textbf{42.59} & \textbf{32.87} & \textbf{25.79} & \textbf{55.72}\\
     \midrule
     \rowcolor[gray]{.8} \multicolumn{11}{c}{Gloss-free} \\
    \midrule
    SLRT$^\dag$ \cite{camgoz2020sign} & 21.03 & 9.97 & 5.96 & 4.04 & 20.51 & 20.00 & 9.11 & 4.93 & 3.03 & 19.67 \\
    GASLT~\cite{yin2023gloss}  & - & - & - & -& - & 19.90  & 9.94& 5.98 & 4.07 &  20.35 \\
    NSLT+Luong \cite{camgoz2018neural,luong2015effective} & 34.22 & 19.72 & 12.24 & 7.96 & 34.28 & 34.16 & 19.57 & 11.84 & 7.56 & 34.54 \\
    \midrule
     \textbf{GFSLT (ours)} &  {37.60} &  {23.30} &  {14.89} &  {9.92} &  {35.42} &  {37.69}  &  {23.28} &  {14.93} &  {9.88} &  {35.16} \\
    \textbf{GFSLT-VLP (ours)} & \textbf{39.20} & \textbf{25.02} & \textbf{16.35} & \textbf{11.07} & \textbf{36.70} & \textbf{39.37} & \textbf{24.93} & \textbf{16.26} & \textbf{11.00} & \textbf{36.44}\\
    \midrule
    \textcolor[gray]{.4}{Improvement} & \textcolor[gray]{.4}{+4.98} & \textcolor[gray]{.4}{+5.30} & \textcolor[gray]{.4}{+4.11} & \textcolor[gray]{.4}{+3.11} & \textcolor[gray]{.4}{+2.42} & \textcolor[gray]{.4}{+5.21} & \textcolor[gray]{.4}{+5.36} &\textcolor[gray]{.4}{+4.42} & \textcolor[gray]{.4}{+3.44} & \textcolor[gray]{.4}{+1.90}\\
    \bottomrule
  \end{tabular}
  \caption{Experimental results on CSL-Daily dataset. Results of SLRT~\cite{camgoz2020sign} and NSLT+Loung~\cite{camgoz2018neural,luong2015effective} are reproduced by \cite{zhou2021improving}, \dag \ denotes our reproduced result under the gloss-free setting.}
  \label{tab:CSL-Daily}
\end{table*}
\subsection{Gloss-free Sign Language Translation}\label{sec:finetune}
In this section, we present our Gloss-Free SLT (GFSLT) network, which can generate the corresponding sentence $S$ from the given sign video $V$ without any gloss annotation assistance. To achieve this, as illustrated in Figure \ref{fig:transformer}, we utilize Transformer \cite{vaswani2017attention} as the main framework of the model, as it has shown superior performance in Neural Machine Translation (NMT).
Initially, the sign video is passed through the Visual Encoder $\psi^*_{VE}$ pretained in the VLP stage to get the hidden semantic vectors:
 \begin{equation}
    h_{1:M} = \psi^*_{VE}(v_{1:T})
\end{equation}
where $M=T/4$.
Meanwhile, Text Decoder $\psi^*_{TD}$ pretained in the VLP stage takes the corresponding sentence $S=(s_1,..,s_U)$ along with the last encoder hidden state as input to generate one word at a time:
\begin{equation}
    z_{u} = \psi^*_{TD}(s_{1:u-1},h_{1:M})
\end{equation}
where the first word of a sentence is artificially set to a special flag word \textless BOS\textgreater , and the Transformer Decoder will end the generation until the flag word \textless EOS\textgreater .
Finally, we calculate the conditional probability $p(S|V)$ after a Linear and a Softmax layer, and optimize the  whole network by minimizing the video-to-sentence cross-entropy loss:
\begin{equation}
    p(S|V) = \prod \limits_{u=1}^U p(s_u|o_u), \quad o_u = softmax(Wz_u+b)
\end{equation}
\begin{equation}
    \mathcal{L}_{g} = - \log p(S|V)
\end{equation}

\section{Experiments}
\subsection{Datasets and Evaluation Metrics}
\noindent\textbf{Datasets}. We evaluated our proposed method on two widely used SLT datasets: RWTH-PHOENIX-Weather 2014T~\cite{camgoz2018neural} and CSL-Daily~\cite{zhou2021improving}. PHOENIX-2014T contains 8257 parallel German sign language (DGS) videos with German translations from weather forecast programs, split into train, dev, and test sets of sizes 7096, 519, and 642 respectively. The German translations have a vocabulary size of 2887. CSL-Daily focuses on daily topics in Chinese sign language, containing 20654 parallel CSL videos with Chinese translations. The dataset is split into train, dev, and test sets of sizes 18401, 1077, and 1176, respectively, and the Chinese translations have a vocabulary size of 2343.

\begin{table*}[!htp]
    \centering
    \resizebox{1\linewidth}{!}{
    \begin{tabular}{cc|c|cccc|cccc}
    \toprule
     \multirow{2}{*}{\textbf{VLP}} & \multirow{2}{*}{\textbf{Aug-S1}} & \multirow{2}{*}{{\textbf{Aug-S2}}} & \multicolumn{4}{c|}{\textbf{Dev}} & \multicolumn{4}{c}{\textbf{Test}}\\
     & & & \textbf{BLEU-1} & \textbf{BLEU-2} & \textbf{BLEU-3} & \textbf{BLEU-4} & \textbf{BLEU-1} & \textbf{BLEU-2} & \textbf{BLEU-3} & \textbf{BLEU-4} \\
    \midrule
    \X & \X & \X & 41.97 & 31.04 & 24.30 & 19.84 & 41.39 & 31.00 & 24.20 & 19.66\\
    \C & \X & \X & 41.50 & 31.26 &  24.64 &  20.12  & 41.81 & 31.34 & 24.40 & 19.77\\
    \C & \C & \X & 42.19 & 32.49 & 26.28 & 22.05 & 42.09 & 32.01 & 25.60 & 21.23\\
    \X & \X & \C & 41.84 & 30.98 & 24.12 & 19.65 & 40.57 & 29.59 & 22.73 & 18.41\\
    \C & \C & \C & \textbf{44.08} & \textbf{33.56} & \textbf{26.74} & \textbf{22.12} & \textbf{43.71} & \textbf{33.18} & \textbf{26.11} & \textbf{21.44}\\
    \bottomrule
    \end{tabular}
    }
    \caption{Effect of VLP and data augmentation strategies. VLP: Visual-Language Pre-training, Aug-S1: strong data augmentation employed during stage 1 for sign video,  Aug-S2: strong data augmentation employed during stage 2 for sign video.
    }
    \label{tab:components}
\end{table*}
\noindent\textbf{Evaluation Metrics}. Following previous works~\cite{camgoz2020sign, zhou2021improving, chen2022simple, chen2022twostream}, we adopt BLEU~\cite{papineni2002bleu} and ROUGE~\cite{lin2004rouge} to evaluate SLT. Higher BLEU and ROUGE-L indicate better translation performance.

\subsection{Implementation details}\label{sec:imple}

\noindent\textbf{GFSLT Model.}  We used ResNet18~\cite{he2016deep} pre-trained on ImageNet~\cite{deng2009imagenet} as our 2D-CNN. For the temporal blocks, we followed the configuration of~\cite{zhou2021improving}, using a stride size of 1/2 and a kernel size of 5/2 for the Conv1D/Maxpooling layers. Our Transformer encoder and decoder both have 3 layers, with a hidden size of 1024 and a feed-forward size of 4096. Each layer has 8 attention heads, and we set the dropout to 0.1 to avoid overfitting.

\vspace{+0.1cm}
\noindent\textbf{Visual-Language Pretraining.} We conduct respective pre-training tasks on the training sets of the two sign language datasets. The mini-batch size is set to 16 (we use AMP~\cite{baboulin2009accelerating} technology to expand the batch size). The input sequences are first resized into $256\times256$, and then randomly/centrally cropped into $224\times224$ during training/inference. We employ SGD as the optimizer and the learning rate is decayed with a cosine schedule~\cite{loshchilov2016sgdr} from 0.01 (maximum) to 1e-5 (minimum). The training lasts for 80 epochs.

\vspace{+0.1cm}
\noindent\textbf{SLT Training and Inference.}
The GFSLT network is trained end-to-end using cross-entropy loss with label smoothing of 0.2 and a mini-batch size of 8. We used SGD optimizer~\cite{robbins1951stochastic} with 0.9 momentum and initialized the learning rate to 0.01 with the cosine annealing scheduler. The network is trained for 200 epochs. During inference, decoding is performed using the beam search strategy with a length penalty~\cite{wu2016google} of 1, and a beam size of 5 is employed.
\begin{table*}[h!]
    \centering
    \resizebox{1\linewidth}{!}{
    \begin{tabular}{cc|c|cccc|cccc}
    \toprule
    \multicolumn{2}{c|}{\textbf{Visual Encoder}} & \multirow{2}{*}{\textbf{T-Decoder}} & \multicolumn{4}{c|}{\textbf{Dev}} & \multicolumn{4}{c}{\textbf{Test}} \\
     \textbf{V-Embedding} &  \textbf{T-Encoder} &  & \textbf{BLEU-1} & \textbf{BLEU-2} & \textbf{BLEU-3} & \textbf{BLEU-4} & \textbf{BLEU-1} & \textbf{BLEU-2} & \textbf{BLEU-3} & \textbf{BLEU-4} \\
    \midrule
    \X & \X & \X & 41.97 & 31.04 & 24.30 & 19.84 & 41.39 & 31.00 & 24.20 & 19.66\\
    \C & \X & \X & 41.31  & 30.88 & 24.25 & 19.83 & 40.12 & 30.03 & 23.34 & 18.93 \\
    \X & \C & \X & 42.75 & 32.31 & 25.65 & 21.23 & 42.94 & 32.68 & 25.83 & 21.25\\
    \C & \C & \X & 43.30 & 32.52 & 26.31 & 22.07 & 43.29 & 32.74 & 25.96 & 21.43\\
    \midrule
     \X & \X & \C & 42.27 & 31.68 & 25.14 & 20.70 & 41.55 & 31.11 & 24.56 & 20.27\\
    \C & \C & \C & \textbf{44.08} & \textbf{33.56} & \textbf{26.74} & \textbf{22.12} & \textbf{43.71} & \textbf{33.18} & \textbf{26.11} & \textbf{21.44} \\
    \bottomrule
    \end{tabular}
    }
    \caption{Investigating the impact of fine-tuning individual components within the Visual-Language-Pretrain (VLP) framework. V-Embedding: Visual Embedding module; T-Encoder: Transformer Encoder; T-Decoder: Transformer Decoder. Notations \X and \C denote the initialization of corresponding layers with random parameters and pre-trained parameters, respectively.
    }
    \label{tab:various}
\end{table*}
\subsection{Comparison with State-of-the-art Methods}
\noindent\textbf{Results on PHOENIX14T dataset.} Table \ref{tab:PHOENIX14T} presents a comparison of our approach with state-of-the-art gloss-based and gloss-free methods for sign language translation. Our method achieves a significant performance gain when compared to other gloss-free approaches, such as CSGCR \cite{zhao2021conditional}. Specifically, our method improves the BLEU-4 score by approximately $+7.0$ and $+5.7$ on the Dev and Test sets, respectively, and improves the ROUGE score by about $+4.8$ and $+2.6$.
Moreover, our results are highly competitive when compared to most gloss-based methods. Notably, our method achieves competitive performance with SLRT \cite{camgoz2020sign} (22.38 $vs.$ 22.12) and STMC-T \cite{zhou2021spatial} (24.09 $vs.$ 22.12), highlighting its potential.


\vspace{+0.1cm}
\noindent\textbf{Results on CSL-Daily dataset.} Table \ref{tab:CSL-Daily} compares our method with the state-of-the-art approaches on the CSL-Daily dataset. CSL-Daily is a large Chinese sign language dataset released in 2021 by \cite{zhou2021improving} and there are therefore only a few methods that tested on it, especially gloss-free ones. Note that the result of SLRT\cite{camgoz2020sign} and NSLT+Loung\cite{camgoz2018neural,luong2015effective} are reproduced by \cite{zhou2021improving}. As it can be seen, our method surpasses the gloss-free method NSLT+Loung\cite{camgoz2018neural,luong2015effective} in all metrics, especially improving the BLEU-4 score about $3.2_{\pm0.1}$ and ROUGE score about $2.2_{\pm0.2}$ on this dataset. Furthermore, compared with gloss-based methods, we are close to the SLRT\cite{camgoz2020sign} and BN-TIN-Transf\cite{zhou2021improving} which did not use semi-supervised back-translation auxiliary training unlike BN-TIN-Transf+SignBT\cite{zhou2021improving}, large model transfer training such as MMTLB\cite{chen2022simple} or a multi-stream model like the one from TS-SLT\cite{chen2022twostream}.

\subsection{Ablation Studies}
The ablation studies were conducted mainly on the PHOENIX14T dataset, with a primary focus on improving the BLEU-4 score as it is the most reliable measure of SLT accuracy. Additionally, unless stated otherwise, we utilized the configuration outlined in Section \ref{sec:imple} as the baseline settings for our network.

\vspace{+0.1cm}
\noindent\textbf{Visual-Language Pretraining.}
In our investigation of VLP, we delved into the key factors that affect its efficacy and discovered several phenomena. Firstly, from Table \ref{tab:components}, we observed that data augmentation on sign videos plays a significant role in the success of VLP. Specifically, when utilizing lightweight data augmentation such as random cropping, the improvement of VLP for SLT is limited, only increasing the BLEU-4 score by about $+0.3$ on the Dev set and $+0.1$ on the Test set. However, when combined with strong data augmentation, VLP significantly enhances the SLT task, improving the BLEU-4 score from 19.84 to 22.05 (+2.2). This emphasizes the data-hungry nature of VLP.
Secondly, we observed that without VLP, relying solely on strong data augmentation in stage 2 does not provide much benefit for SLT and may even impair performance slightly. But when combined with VLP, SLT performance can be continuously improved.
This is because aggressive data augmentation methods may introduce excessive variations or distortions to the training data, which may pose challenges for the SLT model to adapt to the distribution of the augmented data. However, the VLP stage leverages the LLM to encourage the Visual Encoder to adapt to the distribution differences introduced by the augmented data, helping the downstream SLT model develop the ability to generalize from the augmented data.
Additionally, from Table \ref{tab:various}, we find that fine-tuning the Visual Embedding module and Transformer Encoder as a unified whole can result in significant performance gains (+2.23) compared to fine-tuning them separately (-0.01 and +1.39, respectively). Finally, we observe that fine-tuning the Text Decoder can also bring some gains, but it seems limited ($\le 1$). These results confirm that a good visual feature is critical to Gloss-Free SLT. The VLP strategy facilitates the learning of low-redundancy and high-abstract features present in language representations by the Visual Encoder, which makes it a crucial component of the system.



\begin{table}[!t]
    \centering
    \resizebox{1\linewidth}{!}{
    \begin{tabular}{c|cccc}
        \textbf{period} & \textbf{40 epoch} & \textbf{80 epoch} & \textbf{160 epoch} & \textbf{200 epoch}\\
        \toprule
         \rowcolor[gray]{.9} \multicolumn{5}{c}{fixing the training time (80 epochs) of stage 1.} \\
        \midrule
         stage 2 & 18.23 & 20.42 & 21.13 & \textbf{22.12}\\
        \midrule
        \rowcolor[gray]{.9} \multicolumn{5}{c}{fixing the training time (200 epochs) of stage 2.} \\
        \midrule
        stage 1 & 20.62 & 22.12  & \textbf{22.13} & 22.07 \\
    \end{tabular}
    }
    \caption{Effect of longer training regimes. We explore the optimal training time for both stages by fixing the training time of stage 1 to investigate the optimal training time of stage 2 and vice versa in this table.
    }
    \label{tab:pretrain_time}
\end{table}

\vspace{+0.1cm}
\noindent\textbf{Investigation of Training Time.} The training time of gloss-based SLT models typically does not exceed 100 epochs. However, as shown in Table \ref{tab:pretrain_time}, with a fixed pre-training time, the gloss-free SLT model requires a longer training regime ($> 100$ epochs) to achieve satisfactory performance. This is because without the aid of intermediate representation, the convergence speed of the network is reduced, necessitating more training time to make the model fit the desired effect.
Moreover, we investigated the influence of pre-training duration on model performance. As observed, it doesn't seem necessary to have an extended pre-training duration. 80 epochs appears to be a trade-off between the two sign language datasets.


\section{Qualitative Results}
\begin{table}[!htp]
    \centering
    \resizebox{1\linewidth}{!}{
    \begin{tabular}{ll}
        \midrule
        \textbf{Reference:} & sonst regnet es teilweise kräftig\\
                            & (Otherwise it rains heavily at times)\\
        \textbf{GFSLT:} & \colorbox[RGB]{204,255,153}{sonst regnet es} \colorbox[RGB]{255,153,153}{hier und da}\\
                            & (\colorbox[RGB]{204,255,153}{Otherwise it rains} \colorbox[RGB]{255,153,153}{here and there})\\
        \textbf{GFSLT-VLP:} & \colorbox[RGB]{204,255,153}{sonst regnet es teilweise kräftig}\\
                            & (\colorbox[RGB]{204,255,153}{Otherwise it rains heavily at times})\\
        \midrule
        \textbf{Reference:} & am tag wechseln sonne und wolken einander ab
 teilweise ist es auch längere zeit sonnig\\
                            & (During the day sun and clouds alternate partly it is sunny for a long time)\\
        \textbf{GFSLT:} &  \colorbox[RGB]{255,255,153}{am tag sonne und wolken} \colorbox[RGB]{255,153,153}{im wechsel nur vereinzelt schauer}\\
                            & (\colorbox[RGB]{255,255,153}{During the day sun and clouds alternate} \colorbox[RGB]{255,153,153}{only sporadic showers})\\
        \textbf{GFSLT-VLP:} & \colorbox[RGB]{204,255,153}{am tag wechseln sonne und wolken einander ab} \colorbox[RGB]{255,255,153}{es bilden sich längere zeit viel sonnenschein}\\
                            & (\colorbox[RGB]{204,255,153}{During the day, sun and clouds alternate} \colorbox[RGB]{255,255,153}{There is a lot of sunshine for a long time})\\
        \midrule
        \textbf{Reference:} & am tag nur hier und da einige sonnige momente vor allem an den alpen\\
                            & (During the day only here and there some sunny moments, especially in the Alps)\\
        \textbf{GFSLT:} & \colorbox[RGB]{255,255,153}{gebietsweise zeigt sich} \colorbox[RGB]{255,153,153}{morgen} \colorbox[RGB]{255,255,153}{häufig die sonne}\\
                            & (\colorbox[RGB]{255,255,153}{In some areas, the sun will often show up} \colorbox[RGB]{255,153,153}{tomorrow})\\
        \textbf{GFSLT-VLP:} & \colorbox[RGB]{255,153,153}{morgen} \colorbox[RGB]{255,255,153}{zeigt sich mal die sonne wenn dann vor} \colorbox[RGB]{204,255,153}{allem an den alpen}\\
                            & (\colorbox[RGB]{255,153,153}{Tomorrow} \colorbox[RGB]{255,255,153}{the sun will show up}, \colorbox[RGB]{204,255,153}{especially in the Alps})\\
        \midrule
        \textbf{Reference:} & und nun die wettervorhersage für morgen sonntag den zwölften dezember\\
                            & (And now the weather forecast for tomorrow Sunday the twelfth of December)\\
        \textbf{GFSLT:} & \colorbox[RGB]{204,255,153}{und nun die wettervorhersage für morgen sonntag den zwölften} \colorbox[RGB]{255,153,153}{november}\\
                            & (\colorbox[RGB]{204,255,153}{And now the weather forecast for tomorrow Sunday the twelfth of} \colorbox[RGB]{255,153,153}{November})\\
        \textbf{GFSLT-VLP:} & \colorbox[RGB]{204,255,153}{und nun die wettervorhersage für morgen sonntag den zwölften dezember}\\
                            & (\colorbox[RGB]{204,255,153}{And now the weather forecast for tomorrow Sunday the twelfth of December})\\
        \midrule
    \end{tabular}
    }
    \caption{Qualitative results of PHOENIX14T. We highlight the difference between sentences. \colorbox[RGB]{204,255,153}{Green} means totally same as the reference. \colorbox[RGB]{255,255,153}{Yellow} means correct but different words. \colorbox[RGB]{255,153,153}{Red} means totally wrong.}
    \label{tab:Qualitative}
\end{table}
We visually demonstrate our model's performance on several sign language videos from PHOENIX14T test set in Table \ref{tab:Qualitative}. While both models understand the general meaning of sign language videos and produce complete sentences, the baseline model is more error-prone on some keywords, resulting in drastically different translations (first and second rows). Additionally, the VLP model outperforms the baseline in recognizing named entities, accurately translating place names and months (third and fourth rows).

\section{Conclusion and Future work}
In this work, we propose a new perspective for the gloss-free SLT task by reducing the semantic gap between visual and textual representations, which enables us to learn language-indicated visual representations from sign videos. To achieve this, we introduce a novel pre-training paradigm that combines masked self-supervised learning with visual-language supervision learning. Our experiments reveal that both data scale and model parameters have a significant impact on the performance of this method. While our proposed pre-training paradigm is a crucial step towards gloss-free SLT, we acknowledge that further research is needed, especially in pre-training on a large-scale SLT dataset (without gloss annotations). We hope that our work will inspire future research in this area.

\begin{center}
    \section*{Acknowledgement}
\end{center}
This work was supported by the National Key Research and Development Plan under Grant 2021YFE0205700, the External cooperation key project of Chinese Academy Sciences 173211KYSB20200002, the Science and Technology Development Fund of Macau Project 0123/2022/A3, 0070/2020/AMJ, InnoHK program, and has been partially supported by the Spanish project PID2022-136436NB-I00 and by ICREA under the ICREA Academia programme.

{\small
\bibliographystyle{ieee_fullname}
\bibliography{egbib}

\begin{thebibliography}{10}\itemsep=-1pt

\bibitem{baboulin2009accelerating}
Marc Baboulin, Alfredo Buttari, Jack Dongarra, Jakub Kurzak, Julie Langou,
  Julien Langou, Piotr Luszczek, and Stanimire Tomov.
\newblock Accelerating scientific computations with mixed precision algorithms.
\newblock {\em Computer Physics Communications}, 180(12):2526--2533, 2009.

\bibitem{bahdanau2015neural}
Dzmitry Bahdanau, Kyunghyun Cho, and Yoshua Bengio.
\newblock Neural machine translation by jointly learning to align and
  translate.
\newblock In {\em International Conference on Learning Representations}, 2015.

\bibitem{camgoz2018neural}
Necati~Cihan Camgoz, Simon Hadfield, Oscar Koller, Hermann Ney, and Richard
  Bowden.
\newblock Neural sign language translation.
\newblock In {\em Proceedings of the IEEE conference on computer vision and
  pattern recognition}, pages 7784--7793, 2018.

\bibitem{camgoz2020sign}
Necati~Cihan Camgoz, Oscar Koller, Simon Hadfield, and Richard Bowden.
\newblock Sign language transformers: Joint end-to-end sign language
  recognition and translation.
\newblock In {\em Proceedings of the IEEE/CVF conference on computer vision and
  pattern recognition}, pages 10023--10033, 2020.

\bibitem{chen2022simple}
Yutong Chen, Fangyun Wei, Xiao Sun, Zhirong Wu, and Stephen Lin.
\newblock A simple multi-modality transfer learning baseline for sign language
  translation.
\newblock In {\em Proceedings of the IEEE/CVF Conference on Computer Vision and
  Pattern Recognition}, pages 5120--5130, 2022.

\bibitem{chen2022twostream}
Yutong Chen, Ronglai Zuo, Fangyun Wei, Yu Wu, Shujie LIU, and Brian Mak.
\newblock Two-stream network for sign language recognition and translation.
\newblock In Alice~H. Oh, Alekh Agarwal, Danielle Belgrave, and Kyunghyun Cho,
  editors, {\em Advances in Neural Information Processing Systems}, 2022.

\bibitem{vidaug}
darpa-sail on.
\newblock Video augmentation techniques for deep learning.
\newblock \url{https://github.com/darpa-sail-on/videoaug}, 2021.

\bibitem{deng2009imagenet}
Jia Deng, Wei Dong, Richard Socher, Li-Jia Li, Kai Li, and Li Fei-Fei.
\newblock Imagenet: A large-scale hierarchical image database.
\newblock In {\em 2009 IEEE conference on computer vision and pattern
  recognition}, pages 248--255. Ieee, 2009.

\bibitem{devlin2018bert}
Jacob Devlin, Ming-Wei Chang, Kenton Lee, and Kristina Toutanova.
\newblock Bert: Pre-training of deep bidirectional transformers for language
  understanding.
\newblock {\em arXiv preprint arXiv:1810.04805}, 2018.

\bibitem{junwan2016pami}
Sergio Escalera, Jordi Gonz{\`{a}}lez, Xavier Bar{\'{o}}, and Jamie Shotton.
\newblock Guest editors' introduction to the special issue on multimodal human
  pose recovery and behavior analysis.
\newblock 38(8):1489--1491, 2016.

\bibitem{graves2006connectionist}
Alex Graves, Santiago Fern{\'a}ndez, Faustino Gomez, and J{\"u}rgen
  Schmidhuber.
\newblock Connectionist temporal classification: labelling unsegmented sequence
  data with recurrent neural networks.
\newblock In {\em Proceedings of the 23rd international conference on Machine
  learning}, pages 369--376, 2006.

\bibitem{hao2021self}
Aiming Hao, Yuecong Min, and Xilin Chen.
\newblock Self-mutual distillation learning for continuous sign language
  recognition.
\newblock In {\em Proceedings of the IEEE/CVF International Conference on
  Computer Vision}, pages 11303--11312, 2021.

\bibitem{he2016deep}
Kaiming He, Xiangyu Zhang, Shaoqing Ren, and Jian Sun.
\newblock Deep residual learning for image recognition.
\newblock In {\em Proceedings of the IEEE conference on computer vision and
  pattern recognition}, pages 770--778, 2016.

\bibitem{hu2022temporal}
Lianyu Hu, Liqing Gao, Zekang Liu, and Wei Feng.
\newblock Temporal lift pooling for continuous sign language recognition.
\newblock In {\em Computer Vision--ECCV 2022: 17th European Conference, Tel
  Aviv, Israel, October 23--27, 2022, Proceedings, Part XXXV}, pages 511--527.
  Springer, 2022.

\bibitem{imashev2020dataset}
Alfarabi Imashev, Medet Mukushev, Vadim Kimmelman, and Anara Sandygulova.
\newblock A dataset for linguistic understanding, visual evaluation, and
  recognition of sign languages: The k-rsl.
\newblock In {\em Proceedings of the 24th Conference on Computational Natural
  Language Learning}, pages 631--640, 2020.

\bibitem{joze2018ms}
Hamid Reza~Vaezi Joze and Oscar Koller.
\newblock Ms-asl: A large-scale data set and benchmark for understanding
  american sign language.
\newblock In {\em British Machine Vision Conference}, 2019.

\bibitem{kendall2018multi}
Alex Kendall, Yarin Gal, and Roberto Cipolla.
\newblock Multi-task learning using uncertainty to weigh losses for scene
  geometry and semantics.
\newblock In {\em Proceedings of the IEEE conference on computer vision and
  pattern recognition}, pages 7482--7491, 2018.

\bibitem{koller2019weakly}
Oscar Koller, Necati~Cihan Camgoz, Hermann Ney, and Richard Bowden.
\newblock Weakly supervised learning with multi-stream cnn-lstm-hmms to
  discover sequential parallelism in sign language videos.
\newblock {\em IEEE transactions on pattern analysis and machine intelligence},
  42(9):2306--2320, 2019.

\bibitem{konevcny2014one}
Jakub Kone{\v{c}}n{\`y} and Michal Hagara.
\newblock One-shot-learning gesture recognition using hog-hof features.
\newblock {\em The Journal of Machine Learning Research}, 15(1):2513--2532,
  2014.

\bibitem{kornblith2019better}
Simon Kornblith, Jonathon Shlens, and Quoc~V Le.
\newblock Do better imagenet models transfer better?
\newblock In {\em Proceedings of the IEEE/CVF conference on computer vision and
  pattern recognition}, pages 2661--2671, 2019.

\bibitem{li2020word}
Dongxu Li, Cristian Rodriguez, Xin Yu, and Hongdong Li.
\newblock Word-level deep sign language recognition from video: A new
  large-scale dataset and methods comparison.
\newblock In {\em Proceedings of the IEEE/CVF winter conference on applications
  of computer vision}, pages 1459--1469, 2020.

\bibitem{li2020tspnet}
Dongxu Li, Chenchen Xu, Xin Yu, Kaihao Zhang, Benjamin Swift, Hanna Suominen,
  and Hongdong Li.
\newblock Tspnet: Hierarchical feature learning via temporal semantic pyramid
  for sign language translation.
\newblock {\em Advances in Neural Information Processing Systems},
  33:12034--12045, 2020.

\bibitem{li2020transferring}
Dongxu Li, Xin Yu, Chenchen Xu, Lars Petersson, and Hongdong Li.
\newblock Transferring cross-domain knowledge for video sign language
  recognition.
\newblock In {\em Proceedings of the IEEE/CVF Conference on Computer Vision and
  Pattern Recognition}, pages 6205--6214, 2020.

\bibitem{lin2004rouge}
Chin-Yew Lin.
\newblock Rouge: A package for automatic evaluation of summaries.
\newblock In {\em Text summarization branches out}, pages 74--81, 2004.

\bibitem{liu2020multilingual}
Yinhan Liu, Jiatao Gu, Naman Goyal, Xian Li, Sergey Edunov, Marjan
  Ghazvininejad, Mike Lewis, and Luke Zettlemoyer.
\newblock Multilingual denoising pre-training for neural machine translation.
\newblock {\em Transactions of the Association for Computational Linguistics},
  8:726--742, 2020.

\bibitem{loshchilov2016sgdr}
Ilya Loshchilov and Frank Hutter.
\newblock Sgdr: Stochastic gradient descent with warm restarts.
\newblock {\em arXiv preprint arXiv:1608.03983}, 2016.

\bibitem{luong2015effective}
Minh-Thang Luong, Hieu Pham, and Christopher~D Manning.
\newblock Effective approaches to attention-based neural machine translation.
\newblock In {\em Conference on Empirical Methods in Natural Language
  Processing}, pages 1412--1421, 2015.

\bibitem{min2021visual}
Yuecong Min, Aiming Hao, Xiujuan Chai, and Xilin Chen.
\newblock Visual alignment constraint for continuous sign language recognition.
\newblock In {\em Proceedings of the IEEE/CVF International Conference on
  Computer Vision}, pages 11542--11551, 2021.

\bibitem{papineni2002bleu}
Kishore Papineni, Salim Roukos, Todd Ward, and Wei-Jing Zhu.
\newblock Bleu: a method for automatic evaluation of machine translation.
\newblock In {\em Proceedings of the 40th annual meeting of the Association for
  Computational Linguistics}, pages 311--318, 2002.

\bibitem{radford2021learning}
Alec Radford, Jong~Wook Kim, Chris Hallacy, Aditya Ramesh, Gabriel Goh,
  Sandhini Agarwal, Girish Sastry, Amanda Askell, Pamela Mishkin, Jack Clark,
  et~al.
\newblock Learning transferable visual models from natural language
  supervision.
\newblock In {\em International Conference on Machine Learning}, pages
  8748--8763. PMLR, 2021.

\bibitem{robbins1951stochastic}
Herbert Robbins and Sutton Monro.
\newblock A stochastic approximation method.
\newblock {\em The annals of mathematical statistics}, pages 400--407, 1951.

\bibitem{vaswani2017attention}
Ashish Vaswani, Noam Shazeer, Niki Parmar, Jakob Uszkoreit, Llion Jones,
  Aidan~N Gomez, {\L}ukasz Kaiser, and Illia Polosukhin.
\newblock Attention is all you need.
\newblock {\em Advances in neural information processing systems}, 30, 2017.

\bibitem{voskou2021stochastic}
Andreas Voskou, Konstantinos~P Panousis, Dimitrios Kosmopoulos, Dimitris~N
  Metaxas, and Sotirios Chatzis.
\newblock Stochastic transformer networks with linear competing units:
  Application to end-to-end sl translation.
\newblock In {\em Proceedings of the IEEE/CVF International Conference on
  Computer Vision}, pages 11946--11955, 2021.

\bibitem{junwan2013JMLR}
Jun Wan, Qiuqi Ruan, Wei Li, and Shuang Deng.
\newblock One-shot learning gesture recognition from {RGB-D} data using bag of
  features.
\newblock 14(1):2549--2582, 2013.

\bibitem{wu2016google}
Yonghui Wu, Mike Schuster, Zhifeng Chen, Quoc~V Le, Mohammad Norouzi, Wolfgang
  Macherey, Maxim Krikun, Yuan Cao, Qin Gao, Klaus Macherey, et~al.
\newblock Google's neural machine translation system: Bridging the gap between
  human and machine translation.
\newblock {\em arXiv preprint arXiv:1609.08144}, 2016.

\bibitem{yin2023gloss}
Aoxiong Yin, Tianyun Zhong, Li Tang, Weike Jin, Tao Jin, and Zhou Zhao.
\newblock Gloss attention for gloss-free sign language translation.
\newblock In {\em Proceedings of the IEEE/CVF Conference on Computer Vision and
  Pattern Recognition}, pages 2551--2562, 2023.

\bibitem{yu2021searching}
Zitong Yu, Benjia Zhou, Jun Wan, Pichao Wang, Haoyu Chen, Xin Liu, Stan~Z Li,
  and Guoying Zhao.
\newblock Searching multi-rate and multi-modal temporal enhanced networks for
  gesture recognition.
\newblock {\em IEEE Transactions on Image Processing}, 2021.

\bibitem{zhao2021conditional}
Jian Zhao, Weizhen Qi, Wengang Zhou, Nan Duan, Ming Zhou, and Houqiang Li.
\newblock Conditional sentence generation and cross-modal reranking for sign
  language translation.
\newblock {\em IEEE Transactions on Multimedia}, 24:2662--2672, 2021.

\bibitem{Zhou_Li_Wan_2021}
Benjia Zhou, Yunan Li, and Jun Wan.
\newblock Regional attention with architecture-rebuilt 3d network for rgb-d
  gesture recognition.
\newblock {\em Proceedings of the AAAI Conference on Artificial Intelligence},
  35(4):3563--3571, May 2021.

\bibitem{zhou2023pami}
Benjia Zhou, Pichao Wang, Jun Wan, Yanyan Liang, and Fan Wang.
\newblock A unified multimodal de- and re-coupling framework for rgb-d motion
  recognition.
\newblock {\em IEEE Transactions on Pattern Analysis and Machine Intelligence},
  pages 1--15, 2023.

\bibitem{Zhou_2022_CVPR}
Benjia Zhou, Pichao Wang, Jun Wan, Yanyan Liang, Fan Wang, Du Zhang, Zhen Lei,
  Hao Li, and Rong Jin.
\newblock Decoupling and recoupling spatiotemporal representation for
  rgb-d-based motion recognition.
\newblock In {\em Proceedings of the IEEE/CVF Conference on Computer Vision and
  Pattern Recognition (CVPR)}, pages 20154--20163, June 2022.

\bibitem{zhou2021improving}
Hao Zhou, Wengang Zhou, Weizhen Qi, Junfu Pu, and Houqiang Li.
\newblock Improving sign language translation with monolingual data by sign
  back-translation.
\newblock In {\em Proceedings of the IEEE/CVF Conference on Computer Vision and
  Pattern Recognition}, pages 1316--1325, 2021.

\bibitem{zhou2021spatial}
Hao Zhou, Wengang Zhou, Yun Zhou, and Houqiang Li.
\newblock Spatial-temporal multi-cue network for sign language recognition and
  translation.
\newblock {\em IEEE Transactions on Multimedia}, 24:768--779, 2021.

\bibitem{zuo2022c2slr}
Ronglai Zuo and Brian Mak.
\newblock C2slr: Consistency-enhanced continuous sign language recognition.
\newblock In {\em Proceedings of the IEEE/CVF Conference on Computer Vision and
  Pattern Recognition}, pages 5131--5140, 2022.

\end{thebibliography}
}

\appendix
\section{More Implementation details}
\begin{table*}[!htp]
    \centering
    \begin{tabular}{c|cc|c}
    \toprule

    Module & Stride & Kernel & Output Size\\
    \midrule
    Sign Input            & -      & -      & $B \times T \times 224 \times 224 \times 3$ \\
    Resnet wo/ fc    & -      & -      & $B \times T \times 512$\\
    Conv1D-BN1D-RELU &  1     & 5      & $B \times T \times 1024$\\
    MaxPooling1D     &  2      & 2      & $B \times T/2 \times 1024$\\
    Conv1D-BN1D-RELU &  1     & 5      & $B \times T/2 \times 1024$\\
    MaxPooling1D     &  2      & 2      &$B \times T/4 \times 1024$\\
    Linear-BN1D-RELU  &     -       &   -     & $B \times T/4 \times 1024$\\
    Transformer Encoder &     -       &   -     & $B \times T/4 \times 1024$\\
    \midrule
    Text Input  & -      & -      & $B \times U$\\
    Word Embedding & -      & -      & $B \times U \times 1024$\\
    Transformer Decoder & -      & -      & $B \times U \times 1024$\\
    FC & -      & -      & $B \times U \times C$\\
    \bottomrule
    \end{tabular}
    \caption{Detailed Gloss-Free SLT(GFSLT) Framework. B means batch size. T means the lengths of the longest input sign video in the batch. U means the lengths of the longest input text in the batch.
    }
    \label{tab:model}
\end{table*}
\textbf{GFSLT Model} \quad Table \ref{tab:model} presents detailed information on the GFSLT model structure and feature sizes for each module. The input sign video, which may have varying lengths, is padded to the longest length and loaded into a batch. After ResNet~\cite{he2016deep} processing without a fully connected (FC) layer, the resulting visual feature has a size of $B \times T \times 512$. Two temporal modules, each consisting of Conv1D-BN1D-RELU-MaxPooling1D, are used to capture the short-term dependencies in the sign video, yielding features of size $B \times T/4 \times 1024$. These features are then passed through an MLP and a Transformer Encoder to prepare for decoding. In the decoder, the text inputs are first padded to a uniform length of $U$ and passed through a Word Embedding Layer to obtain features of size $B \times U \times 1024$. The Transformer Decoder takes the outputs of the Transformer Encoder and the Word Embedding to generate one word at a time, and an FC layer is used to obtain the final prediction word.

\section{More Ablation Studies}
\subsection{Impact of Model Parameter Size.} 
\begin{figure}[!htp]
\centering
\subfloat[BLEU-4 score on Dev data.]{\includegraphics[width=0.5\linewidth]{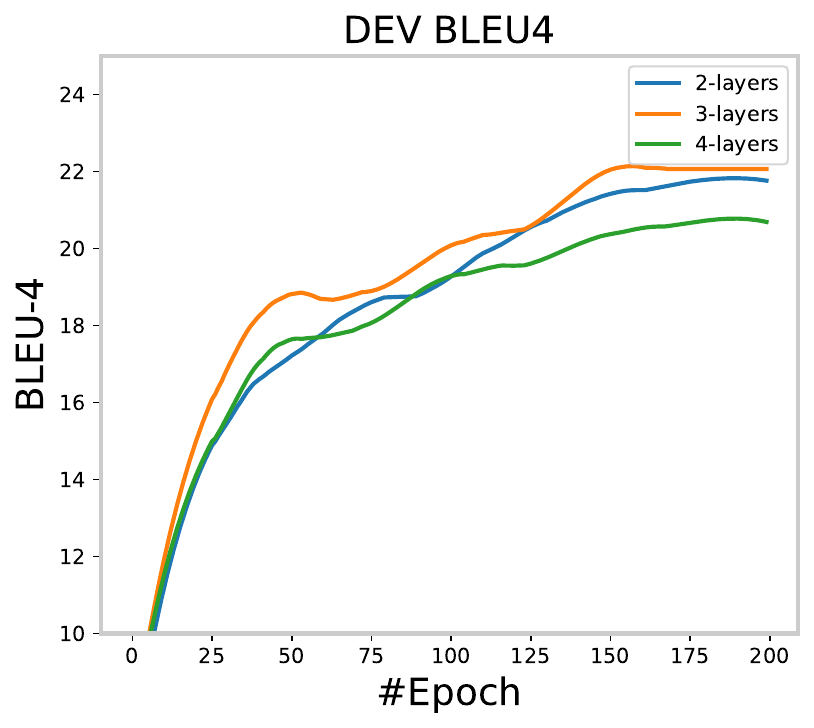}%
\label{subfig:bleu_4}}
\hfil
\subfloat[Loss value on Dev data.]{\includegraphics[width=0.5\linewidth]{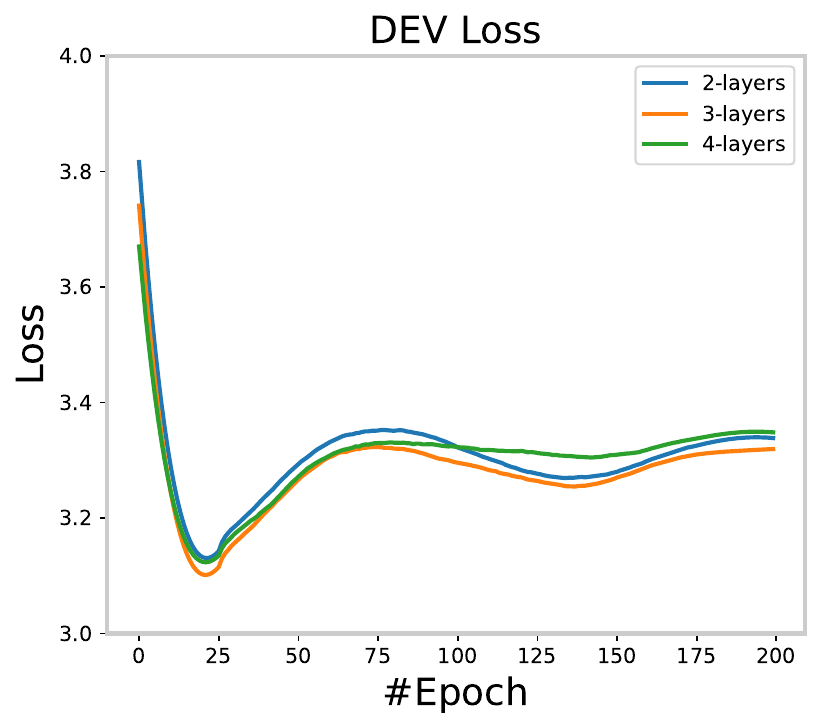}%
\label{subfig:dev_loss}}
\caption{Analysis of the impact of model parameter size. Increasing the network depth (to 4 layers) did not yield any positive results, but instead exacerbated model overfitting.}
\label{fig:depth}
\end{figure}

It is widely acknowledged that the size of the network parameters has a significant effect on the ultimate performance of the model, and a more intuitive perception is that the deeper the network, the better the performance. Nevertheless, for the GFSLT network, we noticed that adding network layers would cause more severe overfitting as shown in Figure \ref{fig:depth}. We attribute this to the limited scale of SLT data, suggesting that a sufficiently large SLT dataset may be able to alleviate this issue.

\subsection{Impact of Mask Rate.}
\begin{table*}[!htp]
    \centering
    \begin{tabular}{c|cccc|cccc}
    \toprule
    \multirow{2}{*}{\textbf{mask rate ($\rho$)}} & \multicolumn{4}{c|}{\textbf{Dev}} & \multicolumn{4}{c}{\textbf{Test}}\\
     & \textbf{BLEU1} & \textbf{BLEU-2} & \textbf{BLEU-3} & \textbf{BLEU-4} & \textbf{BLEU-1} & \textbf{BLEU-2} & \textbf{BLEU-3} & \textbf{BLEU-4} \\
    \midrule
    10\% & 43.30 & 33.05 & 26.04 & 22.03 & 43.54 & 32.90 & 25.61 & 20.84   \\
    \cellcolor[gray]{0.8}15\% & 44.08 & 33.56 & \textbf{26.74} & \textbf{22.12} & 43.71 & \textbf{33.18} & \textbf{26.11} & \textbf{21.44}\\
    20\% & \textbf{44.15} & \textbf{33.72} & 26.35 & 22.07 &  \textbf{43.85} & 33.08 & 25.97 & 21.32 \\
    \bottomrule
    \end{tabular}
    \caption{Effect of mask rate for network performance. The \colorbox[gray]{0.8}{gray box} represents the mask rate we finally adopted in this paper.}
    \label{tab:mask_rate}
\end{table*}
\begin{table*}[!htp]
    \centering
    \resizebox{1\linewidth}{!}{
    \begin{tabular}{cc|cccc|cccc}
    \toprule
    \multirow{2}{*}{\textbf{V-Encoder}} & \multirow{2}{*}{\textbf{T-Encoder}} & \multicolumn{4}{c|}{\textbf{Dev}} & \multicolumn{4}{c}{\textbf{Test}}\\
     & & \textbf{BLEU-1} & \textbf{BLEU-2} & \textbf{BLEU-3} & \textbf{BLEU-4} & \textbf{BLEU-1} & \textbf{BLEU-2} & \textbf{BLEU-3} & \textbf{BLEU-4} \\
    \midrule
    \textit{update} &  \textit{freeze} & 40.28 & 29.83 & 22.13 & 18.32 & 40.81  & 29.32 & 21.24 & 16.93\\
    \cellcolor[gray]{0.8}\textit{update} & \cellcolor[gray]{0.8}\textit{update} & \textbf{44.08} & \textbf{33.56} & \textbf{26.74} & \textbf{22.12} & \textbf{43.71} & \textbf{33.18} & \textbf{26.11} & \textbf{21.44}\\
    \bottomrule
    \end{tabular}
    }
    \caption{Analyze the impact of freezing the Text Encoder during the pretraining stage. \textit{update} means updating the network parameters, and \textit{freeze} means freezing the network parameters. }
    \label{tab:freeze}
\end{table*}

We adopt a token masking strategy in our approach similar to that used in Bert~\cite{devlin2018bert}. Specifically, we randomly replace $\rho$\% of the tokens in a sentence using the following criteria: (i) 80\% of these tokens are replaced with the special [Mask] token, and (ii) 10\% are replaced with any other token, while the remaining 10\% of the tokens are kept intact. As shown in Table~\ref{tab:mask_rate}, our experiments reveal that the optimal BLEU-4 score is achieved with a masking rate of 15\%, which is consistent with the rate used in Bert. Interestingly, we observe that increasing or decreasing the masking rate does not yield significant benefits. This result could be attributed to the fact that the proposed approach, VLP, places more emphasis on pre-training the Visual Encoder than the Text Decoder.

\begin{table*}[!t]
    \centering
    \begin{tabular}{cc|c|cccc|cccc}
    \toprule
     \multirow{2}{*}{\textbf{VLP}} & \multirow{2}{*}{\textbf{Aug-S1}} & \multirow{2}{*}{{\textbf{Aug-S2}}} & \multicolumn{4}{c|}{\textbf{Dev}} & \multicolumn{4}{c}{\textbf{Test}}\\
     & & & \textbf{BLEU-1} & \textbf{BLEU-2} & \textbf{BLEU-3} & \textbf{BLEU-4} & \textbf{BLEU-1} & \textbf{BLEU-2} & \textbf{BLEU-3} & \textbf{BLEU-4} \\
    \midrule
    \X & \X & \X & 37.60 & 23.30 & 14.89 & 9.92 & 37.69 & 23.28 & 14.93 & 9.88\\
    \C & \X & \X &  37.38&    23.26 &  14.91  & 9.97 & 37.84 & 23.60 & 15.23 & 10.29\\
    \C & \C & \X & 38.34 & 24.13 & 15.56 & 10.32 & 38.31 & 23.80 & 15.33 & 10.27\\
    \X & \X & \C & 34.36 & 21.00 & 13.50 & 9.14 & 34.07 & 20.77 & 13.40 & 9.03 \\
    \C & \C & \C & \textbf{39.20} & \textbf{25.02} & \textbf{16.35} & \textbf{11.07} & \textbf{39.37} & \textbf{24.93} & \textbf{16.26} & \textbf{11.00}\\
    \bottomrule
    \end{tabular}
    \caption{Effect of VLP and data augmentation strategies on CSL-Daily dataset. VLP: Visual-Language Pre-training, Aug-S1: strong data augmentation employed during stage 1 for sign video,  Aug-S2: strong data augmentation employed during stage 2 for sign video.
    }
    \label{tab:CSL-Daily}
\end{table*}
\subsection{Impact of Freezing the Text Encoder.}
Considering that the Text Encoder is derived from the pre-trained Mbart~\cite{liu2020multilingual}, so in this experiment, we attempted to freeze its parameters and use it as a teacher model to supervise the learning of the Visual Encoder. Contrary to our expectations, this pre-training strategy did not produce satisfactory results, as shown in Table \ref{tab:freeze}. We hypothesize that the reason for this may be that the text and visual features have fundamentally different underlying representations, and they must be optimized to a common representation for meaningful comparisons and analysis. As a result, directly freezing the parameters of the Text Encoder may not provide sufficient guidance for the Visual Encoder to learn optimal representations in the joint multimodal space.

\subsection{Impact of Loss weight}
\begin{figure}[!htp]
    \centering
    \includegraphics[width=0.8\linewidth]{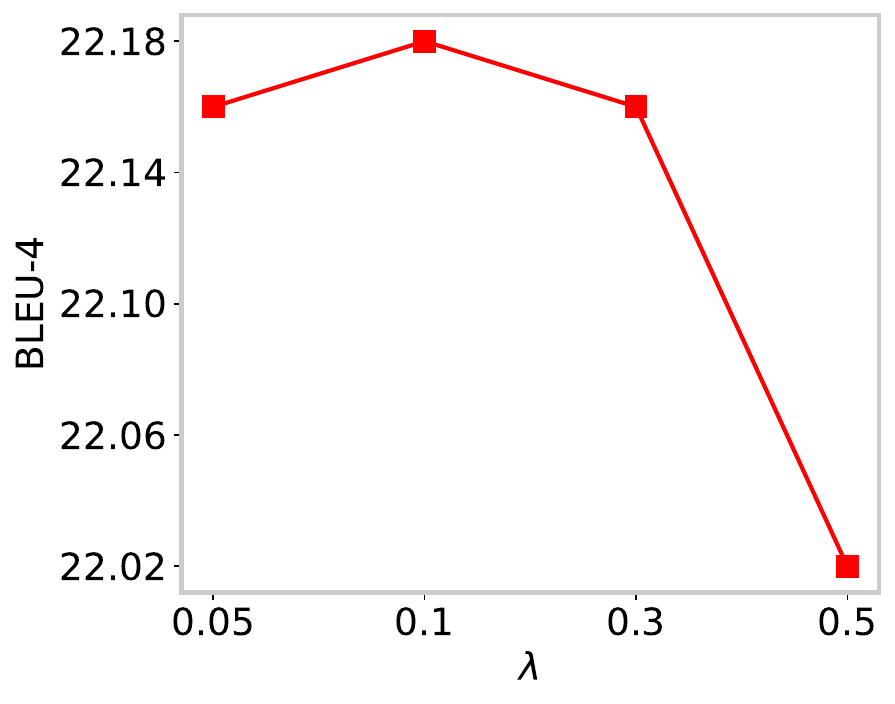}
    \caption{Impact of loss weight coefficient $\lambda$ on network performance.}
    \label{fig:loss_weight}
\end{figure}
In fact, Text Decoder can be updated jointly or in stages. When updating jointly, the loss in the first stage consists of the following two parts:
\begin{equation}
 \mathcal{L}_{total} = \mathcal{L}_s + \lambda \mathcal{L}_c
\end{equation}
where $\lambda$ is a scalar weight.
In this experiment, we studied the effect of the loss weight coefficient $\lambda$ on the pre-trained model. As illustrated in Figure \ref{fig:loss_weight}, the influence of $\lambda$ on the model performance is relatively minor, with performance fluctuations staying around $\pm0.1$. However, as $\lambda$ increases, the model's performance begins to decline, indicating that it is not always beneficial to amplify the influence of the Text Decoder on VLP. As a result, we set $\lambda$ to 0.1 in this paper.

\subsection{Investigation VLP on CSL-Daily}
We also conducted VLP and strong data augmentation ablation experiments on CSL-Daily. As shown in Table \ref{tab:CSL-Daily}, the translation performance improved with VLP, and adding strong data augmentation in Stage 1 further helped. However, the model performance decreased when strong data augmentation was added only in Stage 2 without VLP. The best result was achieved when strong data augmentation was added to both stages. This finding is consistent with the results of our experiments on Phoenix14T.

\end{document}